\documentclass[conference]{IEEEtran}
\IEEEoverridecommandlockouts
\usepackage{cite}
\usepackage{amsmath,amssymb,amsfonts}
\usepackage{algorithmic}
\usepackage{graphicx}
\usepackage{textcomp}
\usepackage{xcolor}
\usepackage{resizegather}

\usepackage{amssymb}
\usepackage{amsmath}
\usepackage{amsthm}
\usepackage[super]{nth}
\usepackage{comment}
\usepackage{graphicx}

\usepackage{subcaption}
\usepackage{bbm}
\usepackage{multirow}
\usepackage{multicol}
\usepackage[ruled,vlined]{algorithm2e}
\usepackage{adjustbox}

\usepackage[ruled,vlined]{algorithm2e}
\usepackage{booktabs}
\usepackage{url}
\usepackage{color, colortbl}
\usepackage{tikz}
\usetikzlibrary{tikzmark, decorations.pathreplacing, calc}
\usepackage{enumitem}
\usepackage{array}
\usepackage{arydshln}
\DeclareMathOperator*{\argmax}{arg\,max}
\DeclareMathOperator*{\argsup}{arg\,sup}
\DeclareMathOperator*{\argmin}{arg\,min}

\DeclareMathOperator*{\KL}{KL}

\DeclareMathOperator*{\Span}{Span}
\DeclareMathOperator*{\supp}{supp}
\newtheorem{theorem}{Theorem}
\newtheorem{lemma}{Lemma}
\newtheorem{definition}{Definition}

\newtheorem{remark}{Remark}

\usepackage{appendix}
\usepackage[numbers,square,sort]{natbib}    
\newcommand{\ep}{\hfill $\Box$}

\def\BibTeX{{\rm B\kern-.05em{\sc i\kern-.025em b}\kern-.08em
    T\kern-.1667em\lower.7ex\hbox{E}\kern-.125emX}}
    
\newcommand\copyrighttext{
\footnotesize \textcopyright 2021 IEEE. Personal use of this material is permitted. Permission from IEEE must be obtained for all other uses, in any current or future media, including reprinting/republishing this material for advertising or promotional purposes, creating new collective works, for resale or redistribution to servers or lists, or reuse of any copyrighted component of this work in other works.}
\newcommand\copyrightnotice{\begin{tikzpicture}[remember picture,overlay]
\node[anchor=south,yshift=10pt] at (current page.south) {\fbox{\parbox{\dimexpr\textwidth-\fboxsep-\fboxrule\relax}{\copyrighttext}}};
\end{tikzpicture}}

\begin{document}
\title{Learning Optimal Antenna Tilt Control Policies:\\ 
A Contextual Linear Bandit Approach 
}

\author{\IEEEauthorblockN{
Filippo Vannella\IEEEauthorrefmark{1}\IEEEauthorrefmark{2}, 
 Alexandre Proutiere\IEEEauthorrefmark{1}, 
 Yassir Jedra\IEEEauthorrefmark{1}, 
 Jaeseong Jeong\IEEEauthorrefmark{2}
\\
\IEEEauthorrefmark{1}KTH Royal Institute of Technology, Stockholm, Sweden\\
\IEEEauthorrefmark{2}Ericsson Research, Stockholm, Sweden\\
Email: \{vannella, alepro, jedra\}@kth.se, jaeseong.jeong@ericsson.com}}

\maketitle
\copyrightnotice

\begin{abstract}
    Controlling antenna tilts in cellular networks is imperative to reach an efficient trade-off between network coverage and capacity. In this paper, we devise algorithms learning optimal tilt control policies from existing data (in the so-called passive learning setting) or from data actively generated by the algorithms (the active learning setting). We formalize the design of such algorithms as a Best Policy Identification (BPI) problem in Contextual Linear Multi-Arm Bandits (CL-MAB). An arm represents an antenna tilt update; the context captures current network conditions; the reward corresponds to an improvement of performance, mixing coverage and capacity; and the objective is to identify, with a given level of confidence, an approximately optimal policy (a function mapping the context to an arm with maximal reward). For CL-MAB in both active and passive learning settings, we derive information-theoretical lower bounds on the number of samples required by any algorithm returning an approximately optimal policy with a given level of certainty, and devise algorithms achieving these fundamental limits. We apply our algorithms to the Remote Electrical Tilt (RET) optimization problem in cellular networks, and show that they can produce optimal tilt update policy using much fewer data samples than naive or existing rule-based learning algorithms.
\end{abstract}

\section{Introduction}
\label{sec:introduction}
In modern wireless communication systems, algorithms for self-optimization of network parameters constitute an essential tool to increase the network efficiency and reduce its operational cost. Important examples of such algorithms are those controlling the Remote Electrical Tilt (RET) of the antennas at Base Stations (BSs). Coverage Capacity Optimization (CCO) is actually one of the key use-cases specified by Self-Organizing Networks (SONs), the network automation technology introduced by the $3^{rd}$ Generation Partnership Project (3GPP) \cite{Waldhauser11}. The goal in CCO is to maximize the network capacity while ensuring that a targeted service area remains covered. There is a clear trade-off between these two performance indicators: tilting down the antenna decreases the coverage at the sector edge and increases the capacity in the sector center. On the other hand, tilting up the antenna may result in excessive sector overlapping; This increases the interference level in surrounding sectors and leads to a high number of unnecessary handovers and ping-pong effects.

In this paper, we investigate the design of algorithms learning tilt control policies achieving an optimal coverage/capacity trade-off. A tilt control policy takes as input the observed network indicators and outputs a tilt update. We consider two scenarios depending on how the data used in the learning process is generated: $(i)$ in the passive learning scenario, this data corresponds to records of previous events in the network; $(ii)$ in the active learning scenario, the policy used to generate data is actively controlled by the learning algorithm (e.g., putting some emphasis on tilt updates for which we are missing information). The passive learning scenario is particularly relevant in real-world applications where there exists a legacy policy deployed into production and there is no possibility to execute unconstrained exploration due to risk or cost reasons. Active learning algorithms can be applied in test-bed network scenarios in which the operator managing the network allows for unconstrained exploration of tilt configurations to determine the best tilt for each sector. 

The objective of the learning algorithms in both the passive and active learning settings is to come up with an approximately optimal tilt update policy using as few data samples as possible. We formalize the design of these algorithms as a Best Policy Identification (BPI) problem in Contextual Linear Multi-Armed Bandit (CL-MAB) \cite{Langford08}. In such a problem, in each round, the learner observes a {\it context}, selects an {\it arm} depending on this context, and receives in return a random {\it reward} whose mean depends on the (context, arm) pair and is initially unknown. Her goal is, using as few samples (or rounds) as possible, to identify a policy that has maximal expected reward with some level of confidence. 

In our case, contexts correspond to current network indicators, arms to the possible tilt updates, and the reward to the resulting improvement in performance, mixing coverage and capacity. We additionally assume that the expected reward can be modeled as an (unknown) linear function of a feature vector encoding a context-action pair \cite{Li11}. This assumption, that we will justify in the case of tilt policy learning, is often made when the number of contexts or arms become large to make the learning problem tractable.

\begin{samepage}
For CL-MAB, in both active and passive learning settings, we derive for the first time information-theoretic lower bounds on the number of samples required by any algorithm returning an approximately optimal policy with a given level of certainty. We devise algorithms in both learning settings achieving these fundamental limits. We then apply our algorithms to the RET optimization problem in cellular networks. Using extensive experiments in a proprietary simulator, we show that our algorithms can produce optimal tilt update policies using much fewer data samples than naive or rule-based algorithms.
\end{samepage}
\section{Previous Work}
\label{sec:previous_work}
\subsection{Contextual Linear Bandits}
Recently, CL-MAB have received considerable attention in the active learning setting for regret minimization \cite{Chu11, Li10, Hao20}. There, the learner aims at minimizing her cumulative regret over a given time horizon. To the best of our knowledge, this paper is the first to analyze and solve the BPI problem in CL-MAB. Closely related work include Best Arm Identification (BAI) in the Linear Bandit (LB) setting without contexts \cite{Soare15, Jedra20, Fiez19}. The BAI problem in the LB setting was initially studied by Soare et al. \cite{Soare15}, that proposes an instance-specific lower bound on the sample complexity as the solution of a max-min optimization problem, and algorithms based on approximations of this optimization problem. Jedra and Proutiere \cite{Jedra20} propose an algorithm whose sampling complexity matches the lower bounds, asymptotically almost surely and in expectation. However, the LB framework is not flexible enough to model more complex applications that require contextual side information. 

\subsection{Antenna Tilt Optimization}
There has been a considerable amount of work in the area of antenna tilt optimization. Recent methods are mainly based on the use of Reinforcement Learning (RL) \cite{Guo13, Balevi19, Vannella21, Dreifuerst21, EricssonMobility}, Contextual Bandits (CBs), \cite{Vannella20} or MABs \cite{Cai10, Gulati14, Dhahri17, Shen18}. Remarkably, RL methods have been actually implemented in real networks and performance gains have been observed \cite{EricssonMobility}. However, the aforementioned papers focus on regret minimization or on identifying an efficient tilt update policy without any consideration for the number of samples used to do so. We should also mention that most existing studies (see e.g, \cite{Vannella20}) investigate off-policy learning problems which correspond to our passive learning setting. Since the proposed methods there do not include any stopping rule, the algorithms may actually stop much before they have collected enough data to learn an optimal policy with reasonable confidence. The way we formulate our BPI problem circumvents this difficulty.  
\section{Problem setting}
\label{sec:problemsetting}
In this section, we introduce various BPI problems in CL-MAB. We then explain how to formalize the RET optimization probem using our contextual bandit framework. 
\subsection{Linear Contextual Bandit - Best Policy Identification}
\label{sec:problem_setting}
In CL-MAB, the decision maker or learner observes a context, selects an action based on this context, and receives a noisy sample of the reward associated to the (context, action) pair. Her objective is to learn as fast as possible the best policy, i.e., the policy selecting an action yielding the maximal expected reward when the context $x$ is observed. We denote by $x_t$ the context observed in round $t$, and assume that it is drawn in an \textit{i.i.d.} (over rounds) manner from $p_\mathcal{X}$, a probability distribution over the context space $\mathcal{X} = \{1,\dots,\mathcal{C}\}\subset \mathbb{R}^d$. The learner sequentially interacts with the environment as follows. Given the observed context $x_t$ and observations made in previous rounds, the learner selects an action from a discrete action set $\mathcal{A}=~\{1,\dots,K\}$. 
In CL-MAB, when the learner selects the action $a_t$, she observes and receives a noisy linear reward $r_t = \theta^\top 
\phi_{x_t,a_t} + \xi_t$, where $\xi_t \sim \mathcal{N}(0,1)$ is the noise, and where the feature map $\phi:\mathcal{X}\times \mathcal{A} \mapsto \mathbb{R}^d$ is known, whereas the vector $\theta \in \mathbb{R}^d$ is unknown. In such CL-MAB, the best policy selects for context $x$ an action $a_\theta^\star(x)\in \arg \max_{a\in \mathcal{A}}\theta^\top\phi_{x,a}$. For any given context-action pair $(x,a)$, we denote by $\nu^\theta_{x,a}~=~\mathcal{N}(\theta^\top \phi_{x,a},1)$ the distribution of the corresponding reward, and by $f_{x,a}^\theta$ its density (w.r.t. the Lebesgue measure). We also define, for any context $x$, $\Phi_x \triangleq \{\phi_{x,a}\}_{a \in \mathcal{A}}$, and $\Phi \triangleq \bigcup_{x\in\mathcal{X}}\Phi_x$. We assume that the dimension of the span of $\Phi_x$ (denoted by $\Span(\Phi_x)$) is $d_x\le d$, and that $\Span(\Phi)=\mathbb{R}^d$. Finally, we denote by $\mathbb{P}_\theta$ the probability measure of the observations generated under the parameter $\theta$, and by $\mathbb{E}_\theta$ the respective expectation.  

In this paper, we investigate the problem of BPI in CL-MAB and in the fixed confidence setting. There, the objective is to devise an algorithm that returns, using as few trials as possible, the best policy (or a nearly optimal policy) with some fixed confidence level. Such an algorithm is defined through a sampling rule, a stopping rule, and a recommendation rule.  
\begin{enumerate}
    \item \textit{Sampling rule:} it specifies the action selected in each round. The sampling rule is a sequence of mappings $(a_t)_{t \ge 1}$, where $a_t: \mathcal{X} \to \mathcal{A}$ may depend on past observations ($a_t(x)$ is the selected action in round $t$ when $x_t=x$). Formally, if $\mathcal{D}_{t-1} = \{(x_s,a_s,r_s)\}_{s = 1}^{t-1}$ represents the history of observations up to round $t$, then $a_t$ is $\mathcal{F}_{t-1}$-measurable, where $\mathcal{F}_{t}$ is the $\sigma$-algebra generated by $\mathcal{D}_{t}$ and the context $x_{t+1}$. 
    \item \textit{Stopping rule:} it controls the end of the data acquisition phase and is defined as a stopping time $\tau$ with respect to the filtration $(\mathcal{F}_t)_{t\ge 1}$ such that $\mathbb{P}_\theta(\tau < \infty) = 1$.
    \item \textit{Recommendation rule:} in round $\tau$, after the data acquisition phase ends, the algorithm returns an estimated best policy $\hat{a}_\tau(x)$, $\forall x \in \mathcal{X}$.
\end{enumerate}

We further distinguish two learning scenarios: $(i)$ in the \textit{active learning} setting, the sampling rule has to be designed, and $(ii)$ in the {\it passive learning} setting, it is fixed and imposed. 

\medskip
\noindent
{\it $(i)$ Active learning.} In this setting, we define an $(\varepsilon,\delta)-$PAC algorithm as follows:
\begin{definition}[\textit{$(\varepsilon,\delta)$-PAC algorithm}] Let $\varepsilon \geq 0$ and $\delta\in(0,1)$. An algorithm is $(\varepsilon,\delta)$-PAC if $\forall \theta \in \mathbb{R}^d$, 
$\mathbb{P}_\theta(\exists x \in \mathcal{X}, \; \theta^\top(\phi_{x,a_{{\theta}}^\star(x)}-\phi_{x,\hat{a}_\tau(x)}) > \varepsilon ) \leq \delta$ and $\mathbb{P}_\theta\left(\tau<\infty\right) = 1$.
\label{def:eps_delta_PAC}
\end{definition}
\noindent The objective for active learning is to devise an $(\varepsilon,\delta)-$PAC algorithm with minimal expected sample complexity $\mathbb{E}_\theta[\tau]$.

\medskip
\noindent
\textit{$(ii)$ Passive learning.} In this setting, the sampling rule is fixed. Specifically, we assume that $a_t(x)$ is selected with probability $\alpha_{x,a}$, independently of the actions selected in previous rounds. Here, an algorithm is defined through its stopping and decision rules only. Note that, since the sampling rule is fixed and imposed, the learner may never have enough information to identify $\theta$ completely, and hence to deduce the best policy. This happens when the set of vectors $\phi_{x,a}$ such that $\alpha_{x,a}>0$ does not span $\mathbb{R}^d$.

\noindent To formalize these cases, we introduce, for any linear subspace $U$ of $\mathbb{R}^d$, the notion of $(U,\varepsilon,\delta)-$PAC algorithm as follows:

\begin{definition}[\textit{$(U,\varepsilon,\delta)$-PAC algorithm}] 
    Let $U$ be a non-empty linear subspace of $\mathbb{R}^d$ with dimension $r\leq d$, $\varepsilon \geq 0$, and $\delta \in (0,1)$. We say that an algorithm is ($U,\varepsilon,\delta$)-PAC if $\forall \theta \in U$, 
    $\mathbb{P}_{{\theta}}\left(\exists x \in \mathcal{X}, \; {\theta}^\top(\phi_{x,a_{{{\theta}}}^\star(x)}-\phi_{x,\hat{a}_\tau(x)}) >\varepsilon\right) \leq \delta$ and $\mathbb{P}_{{\theta}}\left(\tau < \infty\right) = 1$.
    \label{def:u_delta_pac}
\end{definition}

In addition, we say that the linear subspace $U \subset \mathbb{R}^d$ is PAC-learnable if there exists a $(U,\varepsilon,\delta)$-PAC algorithm for some $\varepsilon$ and $\delta$. The objective in the passive learning setting is to devise an $(U,\varepsilon,\delta)-$PAC algorithm with minimal expected sample complexity $\mathbb{E}_\theta[\tau]$ whenever $U$ is PAC-learnable.

\medskip
\noindent
\textbf{Notations.} For a Positive Semi Definite (PSD) matrix $G$ (written $G \succcurlyeq 0$) and a vector $v \in \mathbb{R}^d$, define $\|v\|_G = \sqrt{v^\top Gv}$. Denote by $I_d$ the $d$-dimensional identity matrix. For a matrix $M$, denote by $M^\dagger$ its pseudo-inverse, by $\lambda_i(M)$ its $i$-th smallest eigenvalue, and by $\text{vec}(M)$ its vectorization. Define the Kullback-Leibler (KL) divergence for distributions $\nu$ and $\nu^\prime$ as $\KL(\nu, \nu^\prime)$ and the KL divergence between two Bernoulli distributions of mean $a$ and $b$ as $\mathrm{kl}(a,b)$.  Given two vectors $v$ and $u$, their outer product is denoted by $v\otimes u$. Let $\Lambda = \left\{\alpha \in [0,1]^{\vert \mathcal{X}\vert \times \vert \mathcal{A}\vert}: \forall x\in\mathcal{X} \; \sum_{a\in \mathcal{A}} \alpha_{x,a} = 1 \right\}$ denote the set of vectors representing, for each context, a distribution over arms or actions. For any $\alpha, \alpha' \in\Lambda$, we define $d_\infty(\alpha,\alpha') = \max_{(x,a)\in\mathcal{X}\times \mathcal{A}} |\alpha_{x,a}-\alpha'_{x,a}|$. For $\alpha \in \Lambda$, and a set $C \subseteq \mathbb{R}^{|\mathcal{X}|\times |\mathcal{A}|}$, define $d_\infty(\alpha,C) = \inf_{\alpha' \in C} d_\infty(\alpha,\alpha')$.
 We write $\lesssim$ to denote $\leq$ up to a multiplicative constant. 
\subsection{Remote Electrical Tilt Optimization}
\label{sec:ret_setting}

Consider a mobile network consisting of $\mathcal{S}$ sectors, indexed by $s = 1,\dots,\mathcal{S}$. Each sector $s$ is equipped with an antenna, whose tilt angle at time $t$ is denoted by $v_{s,t}$. We assume there are $\mathcal{U}$ User Equipments (UEs) randomly distributed in the network. Our objective is to control the antenna tilts to maximize the network performance, defined as a function of Key Performance Indicators (KPIs). We are after the best RET policy such that when applied to sectors individually, the network performance is optimized. Devising such a unique best RET policy has the advantage of simplicity and is easier to learn as we can exploit the data from all sectors to learn it. 

\subsubsection{Network performance} We use the framework introduced by Buenostado et al. \cite{Buenostado17} to quantify the network performance. There, the network \textit{coverage} and \textit{capacity} are represented through a set of driver KPIs. These KPIs are sector-specific but include the impact of neighboring sectors. They are functions of the Reference Signal Received Power (RSRP), the received power level from the UEs in the sector, and the Timing Advance (TA) that measures the round-trip time from the serving BS to UEs in a sector. We use the two most significant KPIs, $(i)$ the \textit{sector overshooting factor} and $(ii)$ the \textit{bad coverage} indicators.

The \textit{overshooting factor} $N_{\text{OS}}(s,t)$ for sector $s$ at time $t$ quantifies the interference generated by the antenna of sector $s$ in neighbouring sectors. The set of neighboring sectors is denoted as ${\cal N}(s)$. To construct $N_{\text{OS}}(s,t)$, we look, for a given sector $s'\in {\cal N}(s)$, at the proportion of RSRP samples measured in $s'$ that do not differ significantly (using a fixed threshold for the difference) from the RSRP samples corresponding to the signal coming from the antenna of $s$. This proportion indicates the interference generated by $s$ in $s'$. $N_{\text{OS}}(s,t)$ aggregates these interference indicators for sectors in ${\cal N}(s)$. Refer to \cite{Buenostado17} for a detailed expression of the overshooting factor. 
The \textit{bad coverage} $R_{\text{BC}}(s,t)$ for sector $s$ at time $t$ quantifies the lack of coverage at the sector edge, which can be solved by up-tilting the antenna. An RSRP sample is defined as belonging to the edge of sector $s$ if its TA measurement is beyond the \nth{95} percentile of the TA distribution of sector $s$. $R_{\text{BC}}(s,t)$ is defined as the proportion of edge RSRP samples that are below a given threshold. 

\subsubsection{CL-MAB formulation for RET optimization} \label{sec:LCB_RET_formulation} In CL-MAB, the selected action is a function of the context. Naturally, here, the bad coverage and overshooting indicators provide the context. For a given sector $s$, this context is, at time $t$, $x_{s,t}~=~[1, R_{\text{BC}}(s,t),N_{\text{OS}}(s,t)]~\subseteq~[0,1]^3$. In sector $s$, the action $a_{s,t}$ at time $t$ is chosen from the set $\mathcal{A} = \{[1,0,0],[0,1,0],[0,0,1]\}$, representing tilting the antenna up (\textit{up-tilt}), tilting the antenna down (\textit{down-tilt}) or keeping the same tilt (\textit{no-change}), respectively. The reward obtained for given context and action corresponds to the variations in the two KPIs, weighted by some fixed vector $w=[w_{\text{BC}},w_{\text{OS}}]$ with positive components. At time $t$ in sector $s$, the reward is $p_{s,t} = w_\text{BC}(R_\text{BC}(s,t) -R_\text{BC}(s,t+1)) + w_\text{OS}(N_\text{OS}(s,t) -N_\text{OS}(s,t+1))$. The reward function is the same in all sectors, and we approximate it using a unique linear model $(x,a)\mapsto \phi_{x,a}^\top\theta +\xi$. The feature vector for the (context, action) pair $(x,a)$ is defined as $\phi_{x,a} = \text{vec}\left(x \otimes a \right) \in \mathbb{R}^{9}$ (the vector representation of the outer product of $x$ and $a$). In Section \ref{sec:antenna_tilt_experiments}, we justify the choice of this linear model, and illustrate its accuracy (the model is fitted using the least squares estimator on data from different experiments). 

It is worth emphasizing that the reward model is the same across sectors. Hence, we can use the data gathered in any sector to learn the model $\theta$ and identify the best policy. When applying a BPI algorithm, the successive data samples come from various sectors, and the corresponding contexts can be considered as generated in an \textit{i.i.d.} manner, complying to the CL-MAB framework (refer to Section \ref{sec:experiments} for details).

\section{Sample Complexity Lower Bounds}
We present lower bounds on the expected sample complexity satisfied by any $(\varepsilon,\delta)$-PAC (or $(U,\varepsilon,\delta)$-PAC in the passive learning setting) algorithm in CL-MAB. These bounds extend those derived for the plain linear bandit problems \cite{Soare15} to the case of contextual bandits and to that of identification of $\varepsilon$-optimal policies. Our lower bounds are obtained using classical change-of-measure arguments introduced in \cite{lai1985}, and due to space constraints, we just present a sketch of their proofs. The fact that we wish to identify  an $\varepsilon$-optimal policy (rather than the best policy)  significantly complicates the derivation of the lower bounds; these complications are discussed in details in \cite{garivier2019nonasymptotic}. To state our lower bounds, we introduce the following notations. For $x\in\mathcal{X}$ and $(a,b)\in\mathcal{A}^2: a\neq b$, define $\gamma^x_{a,b} = \phi_{x,b}-\phi_{x,a}$. For any given context $x$, we define the set of $\varepsilon$-optimal arms as $\mathcal{A}_\varepsilon(\theta,x) = \{a\in\mathcal{A}: \theta^\top(\phi_{x,a^\star_\theta(x)}-\phi_{x,a})<\varepsilon\}$, and for $a\notin \mathcal{A}_\varepsilon(\theta,x)$, the set $\mathcal{A}_\varepsilon(\theta,x,a) = \{b\in \mathcal{A}_\varepsilon(\theta,x): \theta^\top \gamma_{a,b}^x \geq \varepsilon\}$. For any $\alpha\in \Lambda$, let  $A(\alpha) = \sum_{(x,a) \in\mathcal{X}\times \mathcal{A}} \phi_{x,a} \phi_{x,a}^\top \alpha_{x,a} p_{\mathcal{X}}(x)$. Finally, we introduce the set $B_\varepsilon(\theta)$ of confusing parameters as $B_\varepsilon(\theta) = {\left\lbrace \mu\in\mathbb{R}^d: \exists x \in \mathcal{X}, \forall a \in \mathcal{A}_\varepsilon(\theta,x), \; \mu^{\top}(\phi_{x,a_{\mu}^\star(x)}- \phi_{x,a})>\varepsilon \right\rbrace}$ (when $\mu\in B_\varepsilon(\theta)$, the best policy for $\mu$ is not $\varepsilon$-optimal for $\theta$). 

\subsection{Active learning}
\begin{theorem}
\label{theorem:lb_eps_delta_discrete}
The sample complexity of any $(\varepsilon,\delta)$-PAC algorithm satisfies, $\forall \theta \in \mathbb{R}^d$, $\mathbb{E}_\theta[\tau] \ge (\psi^\star_{\theta,\varepsilon})^{-1}\mathrm{kl}(\delta,1-\delta)$, where 
\begin{equation}
\label{eq:eps_delta_equivalent_lb}
\psi^\star_{\theta,\varepsilon} = \sup_{\alpha\in \Lambda}\inf_{\mu \in B_\varepsilon(\theta)} \frac{1}{2}\|\mu - \theta\|^2_{A(\alpha)}.
\end{equation}
\end{theorem}

\noindent
{\it Sketch of the proof.} The proof of Theorem \ref{theorem:lb_eps_delta_discrete} proceeds as follows. Consider an $(\varepsilon,\delta)$-PAC algorithm. Let $\mu\in B_\varepsilon(\theta)$. One can show that the expected log-likelihood ratio of the observations under the true parameter $\theta$ and $\mu$ can be written as $\sum_{x,a}\mathbb{E}_\theta \left[N_{x,a}(\tau)\right] \KL(\nu^\theta_{x,a},\nu^{\mu}_{x,a})$ where $N_{x,a}(\tau)$ is the number of times the (context, action) pair $(x,a)$ has been observed before the stopping time $\tau$. Defining $\alpha_{x,a}=\mathbb{E}_\theta \left[N_{x,a}(\tau)\right]/(p_{\cal X}(x)\mathbb{E}_\theta[\tau])$ as the proportion of time arm $a$ is selected when context $x$ appears, we can rewrite the expected log-likelihood ratio as $\frac{\mathbb{E}_\theta[\tau]}{2}\| \theta-\mu\|^2_{A(\alpha)}$. Now define the event $\mathcal{E} = \lbrace \forall x \in \mathcal{X}\; a^\star_\theta(x) = \hat{a}_\tau(x) \rbrace \in \mathcal{F}_\tau$. Since the algorithm is $(\varepsilon,\delta)$-PAC, we have $\mathbb{P}_\theta({\mathcal{E}}) \ge 1- \delta$ and $\mathbb{P}_\mu({\mathcal{E}}) \le\delta$. We conclude, using the data-processing inequality \cite{Garivier19}, that implies $\frac{\mathbb{E}_\theta[\tau]}{2}\| \theta-\mu\|^2_{A(\alpha)}\ge \mathrm{kl}(\mathbb{P}_\theta({\mathcal{E} }),\mathbb{P}_\mu({\mathcal{E} }))\ge \mathrm{kl}(1-\delta,\delta)$. The lower bound is obtained optimizing $\mu$ over the confusing parameters, and selecting the best allocation $\alpha\in \Lambda$.~\ep

\medskip \noindent
{\it Interpretation and tight approximation.} As suggested in the above proof, the allocation $\alpha\in \Lambda$ solving the optimization problem \eqref{eq:eps_delta_equivalent_lb} is optimal: an algorithm relying on a sampling strategy realizing $\alpha$ would yield the lowest possible sample complexity. This allocation however depends on the unknown parameter $\theta$, and to approach it, we need to estimate $\theta$ and repeatedly solve \eqref{eq:eps_delta_equivalent_lb} using these estimators. This is how the so-called track-and-stop algorithms work \cite{Garivier16}. Eq. \eqref{eq:eps_delta_equivalent_lb} is however hard to solve, mainly due to the complexity of the set of confusing parameters $B_\varepsilon(\theta)$.

To circumvent this issue, we propose below a simple and tight approximation of the lower bound, easy to compute and hence easy to leverage in the design of a sampling rule. Specifically, we can show that $T^\star_{\theta,\varepsilon}\mathrm{kl}(\delta,1-\delta)$ with
\begin{equation*}
\label{eq:lb_eps_delta_discrete}
       T^\star_{\theta,\varepsilon} = \inf \limits_{\alpha\in\Lambda} \max\limits_{x\in \mathcal{X}} \max  \limits_{a\notin \mathcal{A}_\varepsilon(\theta,x)}\max\limits_{b\in \mathcal{A}_\varepsilon(\theta,x,a)} \frac{2 \|\gamma^x_{a,b}\|^2_{ A(\alpha)^{-1}}}{ (\theta^\top\gamma^x_{a,b}+\varepsilon)^2},
\end{equation*}
is a tight approximation (actually an upper bound) of the lower bound. It is worth noting that the approximation is exact when the objective is to identify the best policy, i.e., when $\varepsilon=0$. Indeed, adapting the proof of the sample complexity lower bound presented in \cite[Theorem 3.1]{Soare15} to the case of CL-MAB, one can readily show that, for any $(0,\delta)$-PAC (or simply $\delta$-PAC) algorithm, the sample complexity lower bound is $T^\star_\theta \mathrm{kl}(\delta,1-\delta)$, with\\ 
$$ 
    T^\star_\theta = \inf\limits_{\alpha \in \Lambda}\max\limits_{x \in \mathcal{X}, a \in \mathcal{A}\setminus\{a^\star_\theta(x)\}} \frac{\|\phi_{x,a_\theta^\star(x)} - \phi_{x,a}\|^2_{ A(\alpha)^{-1}}}{(\theta^\top
    (\phi_{x,a_\theta^\star(x)} - \phi_{x,a}))^2}.$$
This bound coincides with our approximation for $\varepsilon=0$. 

We note that an equivalent way to define $(T^\star_{\theta,\varepsilon})^{-1}$ is through a convex program: $\bar{\psi}^\star_{\theta,\varepsilon} = \sup_{\alpha\in\Lambda}\bar{\psi}_{\theta,\varepsilon}(\alpha)$, where 
$\bar{\psi}_{\theta,\varepsilon}(\alpha) = \inf_{\mu \in \bar{B}_{\varepsilon}(\theta)} \frac{1}{2}\|\theta - \mu\|^2_{A(\alpha)},$
and $\bar{B}_\varepsilon(\theta) = {\left\lbrace \mu\in\mathbb{R}^d: \exists x \in \mathcal{X}, \exists a \in \mathcal{A}_\varepsilon(\theta,x), \; \mu^{\top}(\phi_{x,a_{\mu}^\star(x)}- \phi_{x,a})>\varepsilon \right\rbrace}$.
The following lemma provides a set of properties $\bar{\psi}_{\theta,\varepsilon}$ satisfies.
\begin{lemma}
\label{lem:properties_eps_delta} Let $C_\varepsilon^\star(\theta) = \argsup_{\alpha \in \Lambda} \bar{\psi}_{\theta,\varepsilon}(\alpha)$. For any $\varepsilon\geq 0$, we have
\begin{enumerate}
    \item $\bar{\psi}_{\theta,\varepsilon}(\alpha)$ is continuous in $\theta$ and $\alpha$,
    \item $\exists \; \alpha^\star \in C^\star_{\varepsilon}(\theta)$ such that $A(\alpha^\star) \succ 0$,
    \item $\bar{\psi}_{\theta,\varepsilon}^\star$ is continuous in $\theta$,
    \item $C_{\varepsilon}^\star(\theta)$ is convex, compact and upper hemi-continuous. 
\end{enumerate}
\end{lemma}
\noindent The proof of the above lemma is similar to the one presented in \cite[Lem. 1 \& 2]{Jedra20}. We finish the active learning section by presenting a lemma that gives an upper bound of $T^\star_{\theta,\varepsilon}$. 
\begin{lemma}
\label{lem:T_star_scaling_eps_delta_pac}
$\forall \theta \in \mathbb{R}^d$, $T^\star_{\theta,\varepsilon} \le \frac{d}{\varepsilon^2} \mathbbm{1}_{ \{ \varepsilon>0\} } +\frac{4d}{\Delta_{\min}^2}\mathbbm{1}_{ \{ \varepsilon = 0\} }$, where $\Delta_{\min}=\min_{x, a\neq a_\theta^\star(x)} \theta^\top (\phi_{x,a_\theta^\star(x)}-\phi_{x,a})$ is the minimal gap between the optimal arm and a sub-optimal arm. 
\end{lemma}

\noindent This lemma is proved using similar techniques as those used in \cite[Lem. 2]{Soare15} for plain linear bandits. It provides a worst-case scaling of our lower bound. Note that the lower bound does not scale with the number of contexts (this is expected as $\theta$ is a unique parameter driving the reward in {\it all} contexts). Importantly, the number of samples required to identify the best policy grows as $1/\Delta_{\min}^2$, and typically the minimal gap $\Delta_{\min}$ goes to 0 as the number of contexts grows large. When applying the CL-MAB framework to the RET optimization problem, we may face a very large number of contexts, and this is why we wish to identify $\varepsilon$-optimal policies. Looking for the best policy instead would require too many samples. 

\subsection{Passive learning}
In the passive learning setting, the way arms are selected is fixed and defined through $\alpha\in \Lambda$. Denote by $\Phi_\alpha=\{\phi_{x,a}: \alpha_{x,a}>0\}$. The following theorem states that if $U$ is included in $\mathrm{Span}(\Phi_\alpha)$, then $U$ is PAC-learnable, in which case it also provides a sample complexity lower bound satisfied by any $(U,\varepsilon,\delta)$-PAC algorithm. When $U$ is not included in $\mathrm{Span}(\Phi_\alpha)$, we can easily build an example where $U$ is not PAC-learnable. Indeed, using the data, we can only learn the projection of $\theta$ on $\mathrm{Span}(\Phi_\alpha)$; now, select two vector $\theta$ and $\mu$ whose projections on $\mathrm{Span}(\Phi_\alpha)$ coincide but with different optimal policy (this happens only if for some context $x$, $\alpha_{x,a_\mu^\star(x)}$ or $\alpha_{x,a_\theta^\star(x)}$ is $0$). Obviously, in this case, we cannot learn the optimal policy from the data.  

\begin{theorem}
\label{theorem:off_policy_lb}
Let $U$ be a linear subspace of $\mathrm{Span}(\Phi_\alpha)$. $U$ is PAC-learnable and the sample complexity of any $(U,\varepsilon,\delta)$-PAC algorithm satisfies, $\forall \theta \in U$, $\mathbb{E}_{{\theta}}[\tau]\ge (\psi_{{\theta,\varepsilon}}(\alpha))^{-1}\mathrm{kl}(\delta,1-\delta)$, where $
\psi_{\theta,\varepsilon}(\alpha)=\inf_{\mu \in B_\varepsilon(\theta)\cap U} \frac{1}{2}\|\mu - \theta\|^2_{A(\alpha)}.$
\end{theorem}
    Note that the maximal PAC-learnable set $U$ is $\mathrm{Span}(\Phi_\alpha)$. As for the active learning setting, we can approximate the above lower bound by $T_{\theta,\varepsilon}(\alpha) \mathrm{kl}(\delta,1-\delta)$, with
\begin{equation*}
        T_{{\theta},\varepsilon}(\alpha) = \max\limits_{x\in \mathcal{X}} \max  \limits_{a\notin \mathcal{A}_\varepsilon({\theta},x)}\max\limits_{b\in \mathcal{A}_\varepsilon({\theta},x,a)} \frac{2 \|P\gamma_{a,b}^x\|^2_{ (PA(\alpha)P^\top)^{\dagger}}}{ ({\theta}^\top\gamma_{a,b}^x+\varepsilon)^2},
\end{equation*}
and where we define $P$ be the $r\times d$  orthogonal matrix such that  $\textrm{rows}(P) = \{u_1^\top,\dots,u_r^\top\}$, and $(u_1,\dots,u_r)$ is an orthonormal basis of the subspace $U$.
\section{Algorithms}
In this section, we present our algorithm for the the active and passive learning settings. The first part of the section (see \ref{sec:LSE}, \ref{sec:stopping_rule}, and \ref{sec:sampling_rule}) describes the algorithm for active learning setting, while \ref{sec:passive_learning_algorithm} presents the algorithm in the passive learning setting. The algorithms apply the track-and-stop framework developed in \cite{Garivier16} and \cite{Jedra20} for regular and LB problems. In the active learning setting, the algorithm consists in $(i)$ estimating the unknown parameter $\theta$ using the Least-Squares Estimator (LSE), $(ii)$ using this estimator to compute the optimal sampling rule suggested by the lower bound optimization problem leading to $T_{\theta,\varepsilon}^\star$, and $(iii)$ tracking this sampling rule and stopping when enough information has been gathered. We detail these steps below. 

\subsection{Least-Squares Estimator}
\label{sec:LSE}
The LSE for $\theta$ can be explicitly computed and is, after gathering $t\ge 1$ samples,
\begin{equation}
        \hat{\theta}_t = \left(\sum_{s = 1}^t \phi_{x_s,a_s}\phi^\top_{x_s,a_s}\right)^\dagger\left(\sum_{s = 1}^t\phi_{x_s,a_s}r_{s}\right).
        \label{eq:LS_estimator}
\end{equation}
As in absence of contexts, the quantity that will control the performance of the LSE is the smallest eigenvalue of the covariates matrix $A_t = \sum_{s = 1}^t \phi_{x_s,a_s}\phi^\top_{x_s,a_s}$. The following results are straightforward extensions of similar results derived for LB without context \cite{Jedra20}, and obtained applying concentration results for self-normalized processes \cite{Abbasi11}. 

\begin{lemma}
\label{lem:LS_concentration}
(i) If the sampling rule satisfies, for some $\alpha\in(0,1)$, $\lim\inf_{t \to \infty }\lambda_{1}\left( \frac{A_t}{t^\alpha} \right) > 0$ a.s., then $\lim_{t \to \infty} \hat{\theta}_t =  \theta$ a.s. In particular, for all $\beta \in (0,\alpha/2)$,
   $ \|\hat{\theta}_t - \theta\| = o\left(t^{-\beta}\right)$ a.s.. (ii)     Let $L = \max_{x,a}\|\phi_{x,a}\|$ and assume there exists $t_0 \geq 1$ such that for all $t\geq t_0$, $\lambda_{1}\left(A_t\right) \geq c t^\alpha$ a.s. for some $c>0$. Then for all $t\geq t_0$,
    $$
    \mathbb{P}_\theta(\|\hat{\theta}_t - \theta\| \geq \varepsilon) \leq \left(c^{-1 / 2} L\right)^{d} t^{\frac{(1-\alpha) d}{2}} \exp \left(-\frac{c \varepsilon^{2} t^{\alpha}}{4}\right).
    $$
\end{lemma}

\subsection{Stopping and decision rule}
\label{sec:stopping_rule}
Both rules leverage the Generalized Log-likelihood Ratio (GLR) $Z^x_{a, b, \varepsilon}(t)$ indicating whether arm $a$ is better than $b$ (up to a precision $\varepsilon$). For any pair of arms $(a,b)\in\mathcal{A}^2:a \neq b$, $x\in\mathcal{X}$, $t\geq1$, and $\varepsilon \geq 0$, it is defined as
{\begin{equation*}
Z^x_{a, b, \varepsilon}(t)=\log \left(\frac{\max_{\left\{\theta\in\mathbb{R}^d: \theta^{\top}\gamma_{a,b}^x \leq \varepsilon\right\}} \prod_{s=1}^t f_{x_s,a_s}^{\theta}(r_s)}{\max_{\left\{\theta \in\mathbb{R}^d:\theta^{\top}\gamma_{a,b}^x \geq \varepsilon \right\}} \prod_{s=1}^t f_{x_s,a_s}^{\theta}(r_s) }\right). 
\end{equation*}}
We can easily obtain an explicit expression of $Z^x_{a,b}$ as in \cite{Jedra20} for plain linear bandits. For $t\geq 1$ such that $A_t \succ 0$, for all $x \in\mathcal{X}$, and $(a,b) \in\mathcal{A}^2: a\neq b$, we have
\begin{equation*}
    Z^x_{a, b, \varepsilon}(t)=\operatorname{sgn}\left(\hat{\theta}_{t}^{\top}\gamma^x_{b,a}+\varepsilon\right) \frac{\left(\hat{\theta}_{t}^{\top}\gamma^x_{b,a}+\varepsilon\right)^{2}}{2\|\gamma^x_{b,a}\|_{A_t^{-1}}}.
\end{equation*}
\noindent
At the stopping time $\tau$, the decision rule is just defined as
\begin{equation}
\label{eq:recommendation_rule}
    \hat{a}_\tau(x) = \argmax_{a \in \mathcal{A}_\varepsilon(\hat{\theta}_\tau,x)} \min_{b \neq a} Z^x_{a, b, \varepsilon}(\tau), \quad \forall x\in\mathcal{X},
\end{equation}
and it can be computed using the above explicit expression of $Z^x_{a, b, \varepsilon}(t)$. The stopping rule is based on the GLR and the classical Chernoff stopping rule \cite{Garivier16}.

Define $Z_\varepsilon^x(t) = \max_{a \in \mathcal{A}_\varepsilon(\hat{\theta}_t,x)} \min_{b \neq a} Z^x_{a, b, \varepsilon}(t)$. The stopping time is 
\begin{align}
\begin{split}
    \tau =\inf \Big\{t:\forall x\in\mathcal{X}, Z_\varepsilon^x(t)>\beta(\delta, t) \text{ and } A_t \succeq c I_{d}\Big\},
\label{eq:stopping_rule_eps_delta_pac}
\end{split}
\end{align}
where $\beta(\delta,t)$ is an exploration threshold chosen to ensure that the algorithm is $(\varepsilon,\delta)$-PAC, and $c$ is a strictly positive constant. An exploration threshold leading to an $(\varepsilon,\delta)$-PAC algorithm is proposed in the following lemma. The proof sketches of all lemmas presented in this section are postponed to Section \ref{sec:lemmas}.
\begin{lemma}
\label{lem:stopping_rule_eps_delta_pac}
Let $u>0$, and define the exploration threshold
\begin{equation}
    \label{eq:expl_rate_eps_delta_pac}
        \beta(\delta, t) = (1+u) \log \left(\frac{\operatorname{det}\left((uc)^{-1}  A_t+I_{d}\right)^{\frac{1}{2}}}{\delta}\right).
\end{equation}
\noindent
Under the stopping rule defined by \eqref{eq:stopping_rule_eps_delta_pac}--\eqref{eq:expl_rate_eps_delta_pac} and under any sampling rule, we have
     $$\mathbb{P}_{\theta}\left(\exists x\in \mathcal{X}: \theta^\top (\phi_{x,a^\star_\theta(x)} - \phi_{x,\hat{a}_\tau(x)}) >\varepsilon, \tau <\infty\right)\leq\delta.$$
\end{lemma}
\noindent Note that the exploration threshold has the nice property of being independent on the number of contexts $|\mathcal{X}|$.

\subsection{Sampling rule} 
\label{sec:sampling_rule}
To obtain an algorithm with minimal sample complexity, we need to devise a sampling rule approaching an optimal proportion of arm draws $\alpha_\theta^\star\in C^\star_{\varepsilon}(\theta)$. To this aim, we propose a sampling rule based on two components: a forced exploration phase ensuring that $\theta$ is estimated accurately, and a tracking phase where, based on the estimated $\theta$, the allocation of arm draws is maintained close to the set $C^\star_{\varepsilon}(\theta)$.

\textit{a) Forced Exploration.} Lemma \ref{lem:LS_concentration} presents conditions under which the LSE of $\theta$ performs well. The forced exploration phase is designed so that these conditions hold. To design the forced exploration phase, we pick, for each context $x$, a set of arms $\mathcal{A}_x = \left\{{a}_x(1),\dots,{a}_x(d_x)\right\} \subseteq \mathcal{A}$ such that $\lambda_{d_x}\left(\sum_{{a}_x \in {\mathcal{A}}_x}\phi_{x,{a}_x}\phi_{x,{a}_x}^\top\right)>0$. The sets are chosen so that there exists $c>0$ such that $\sum_{x\in\mathcal{X}} \sum_{{a}_x \in {\mathcal{A}}_x}\phi_{x,{a}_x}\phi_{x,{a}_x}^\top \succeq c I_d$. Let $A_{t}(x) = \sum_{s = 1}^t\phi_{x_s,a_s}\phi_{x_s,a_s}^\top \mathbbm{1}_{\{x_s=x\}}$, $N_{x}(t) = \sum_{s = 1}^t \mathbbm{1}_{\{x_s=x\}}$, and $p_{\min} = \min\limits_{x\in\mathcal{X}}p_\mathcal{X}(x)$. The forced exploration phase is defined and analyzed in the following lemma. 
\begin{lemma}
\label{lem:forced_exploration}
      Let $(d_t(x))_{t\geq 1}$ be an arbitrary sequence of context-action mappings. for $t \geq 1$ and for all $x \in \mathcal{X}$. Define  $f_x(t) = \sqrt{\frac{N_x(t)}{d_x}} \sum_{{a}_x \in {\mathcal{A}}_x}\phi_{x,{a}_x}\phi_{x,{a}_x}^\top$, and the index $i_t(x)$ such that $i_0(x) = 1$, $i_{t+1}(x) = (i_t(x) \mod d_x) +\mathbbm{1}_{\{A_t(x) \prec f_x(t)\}}$. Define the sampling rule as
\begin{equation}
\label{eq:sampling_rule_force}
    a_{t+1}(x) =\left\{\begin{array}{ll}
    d_{t}(x) & \text {if } \quad A_t(x) \succeq f_x(t) \\
    a_x\left(i_{t}(x)\right)& \text {otherwise } \end{array}\right..
\end{equation}
Then $\forall \varepsilon \in(0,p_{\min})$, there exist $\kappa>0$, $t_1(\kappa)\geq1$ such that $\forall t\geq t_1(\kappa)$, we have
$\mathbb{P}_\theta\left(\frac{1}{\sqrt{t}} A_t \succeq \kappa I_d \right) \geq 1-2e^{\frac{t\varepsilon^2}{4}}$.
\end{lemma}

As described above, in the exploration phase, for the context $x$, we select an arm from $\mathcal{A}_x$ when necessary so that when covering all contexts, we increase the smallest eigenvalue of the covariates matrix. The result in Lemma \ref{lem:forced_exploration} specifying the rate at which this smallest eigenvalue increases is probabilistic and obtained using concentration results on the time taken to cover all contexts (contexts are \textit{i.i.d.} over time). This contrasts with the plain linear bandit scenario, where a similar forced exploration phase \cite{Jedra20} would actually lead to a deterministic growth of this eigenvalue. We show that the probabilistic guarantees of Lemma \ref{lem:forced_exploration} are sufficient to obtain, together with Lemma \ref{lem:LS_concentration} the appropriate convergence of the LSE $\hat{\theta}_t$ to $\theta$. 

\textit{b) Tracking rule.} To complete the design of the sampling rule, it remains to determine the sequence of context-action mappings $(d_t(x))_{t\geq 1}$. Recall that the optimal sampling rule must match an optimal proportion of arm draws in $C_\varepsilon^{\star}(\theta)$. Since $\theta$ is unknown, our sampling rule tracks, at time $t$, an estimated optimal allocation $\alpha(t) \in C_\varepsilon^{\star}(\hat{\theta}_t)$. The following lemma provides a way to devise such tracking rule. 
\begin{lemma}
\label{lem:tracking_lemma}
    Let $(\alpha(t))_{t \geq 1}$ be a sequence taking values in $\Lambda$, such that there exists a compact, convex and non-empty set $C \subset \Lambda$, there exist $\varepsilon>0$ and $t_0(\varepsilon)>0$, such that for all $t\geq t_0(\varepsilon)$, $d_\infty(\alpha(t),C)\leq \varepsilon$. Define 
\small{\begin{equation}
    \label{eq:sampling_rule_track}
    d_t(x) = \argmin_{a \in \supp \left(\sum_{s = 1}^t \alpha(s) \mathbbm{1}_{\{x_s = x\}}\right)} \left( N_{x,a}(t) - \sum_{s = 1}^t \alpha_{x,a}(s) \right), 
\end{equation}}\normalsize
\normalsize where $N_{x,a}(t) = \sum_{s = 1}^t \mathbbm{1}_{\{a_s = a,x_s = x\}}$. 
Then $\forall u \in (0,p_{\min})$, there exist $t_1(\varepsilon,u) \geq t_0(\varepsilon)$ such that $\forall t \geq t_1(\varepsilon,u)$, we have
\footnotesize{$$\mathbb{P}_\theta\left( d_\infty\left(\left(\frac{N_{x,a}(t)}{N_x(t)}\right)_{(x,a)\in\mathcal{X}\times\mathcal{A}},C \right) \le \varepsilon(z_t+d-1)\right)\geq 1-2e^{\frac{t u^2}{4}},$$
where $z_{t} = \max_{x\in \mathcal{X}} |\supp(\sum_{s = 1}^t \alpha(s)\mathbbm{1}_{\{x_s = x\}} ) \setminus \mathcal{A}_x|$.}\normalsize
\end{lemma}

To conclude the description of our sampling rule, we specify the sequence $(\alpha(t))_{t\geq 1}$ (whose asymptotic optimality will come from the previous lemma). We actually only require that 
this sequence satisfies the following condition: there exists $y >0$, and a non-decreasing sequence $(\ell(t))_{t\geq 1}$ such that $\ell(1) = 1$, $\ell(t) \leq t$ and $\lim\inf_{t\to \infty} \ell(t)/t^p $ for some $p>0$ and such that $\forall \kappa>0, \exists h(\kappa): \forall t \geq 1,$
\begin{equation}
    \label{eq:condition_convergence}
     \mathbb{P}\left(\min_{s \geq \ell(t)} d_{\infty}\left(\alpha(t), C_\varepsilon^{\star}\left(\hat{\theta}_{s}\right)\right)>\kappa\right) \leq \frac{h(\kappa)}{t^{2+y}}.
\end{equation}
The condition \eqref{eq:condition_convergence} is easy to ensure in practice (see \cite{Jedra20} for examples of updates of $\alpha(t)$ satisfying such condition).  

The following lemma summarizes the performance of our sampling rule.
\begin{lemma}
\label{lem:convergence_proportions}
Under any sampling rule defined by \eqref{eq:sampling_rule_force}-\eqref{eq:sampling_rule_track} and satisfying condition \eqref{eq:condition_convergence}, we have
$$\lim_{t\to \infty}d_\infty\left(\left(\frac{N_{x,a}(t)}{N_x(t)}\right)_{(x,a)\in\mathcal{X}\times\mathcal{A}},C^\star_\varepsilon(\theta) \right) = 0, \text { a.s.}.$$
\end{lemma}

\subsection{Passive learning algorithm}
\label{sec:passive_learning_algorithm}
We detail below the decision rule and the stopping rule defining our BPI algorithm in the passive learning setting.

\textit{a) Decision rule:} The decision rule is defined in terms of the GLR for parameters $\theta \in U$, expressed as
{\begin{equation*}
Z^x_{U,a,b,\varepsilon}(t) = \log \left(\frac{\max_{\left\{\theta \in U : \theta^{\top}\gamma_{a,b}^x \leq \varepsilon\right\}} \prod_{s=1}^t f_{x_s,a_s}^{\theta}(r_s)}{\max_{\left\{\theta \in U :\theta^{\top}\gamma_{a,b}^x \geq \varepsilon \right\}} \prod_{s=1}^t f_{x_s,a_s}^{\theta}(r_s)}\right).
\end{equation*}}
As for the active setting, we can express $Z^x_{U,a,b,\varepsilon}(t)$ explicitly for $t\geq 1$ such that $P A_t P^\top \succ0$, as 
$$Z^x_{U,a,b,\varepsilon} = \operatorname{sgn}\left(\hat{\theta}_{U,t}^{\top}(P\gamma^x_{b,a})+\varepsilon\right) \frac{\left(\hat{\theta}_{U,t}^{\top}(P\gamma^x_{b,a})+\varepsilon\right)^{2}}{2\|P \gamma^x_{b,a}\|_{(P A_t P^\top)^{-1}}},$$
where $\hat{\theta}_{U,t} = \left(P A_t P^\top\right)^{-1}P \left(\sum_{s = 1}^T \phi_{x_s,a_s}r_s \right)$. Note that $\hat{\theta}_{U,t} \in\mathbb{R}^r$. The \textit{decision rule} is then defined as
\begin{equation}
\label{eq:recommendation_rule_passive}
    \hat{a}_\tau(x) = \argmax_{a \in \mathcal{A}_\varepsilon(\hat{\theta}_{U,\tau},x)} \min_{b \neq a} Z^x_{U,a, b, \varepsilon}(\tau), \quad \forall x\in\mathcal{X}.
\end{equation}

\textit{b) Stopping rule:} Denote by $Z_{U,\varepsilon}^x(t) = \max_{a \in \mathcal{A}_\varepsilon(\hat{\theta}_{U,\tau},x)} \min_{b \neq a} Z^x_{U, a, b, \varepsilon}(t)$. Then, the \textit{stopping rule} is defined as
\small{\begin{align}
    \begin{split}
    \label{eq:stopping_rule_eps_delta_pac_passive}
        \tau =\inf \Big\{t: \forall x, Z_{U,\varepsilon}^x(t)>\beta_U(\delta, t) \text{ and } P A_t P^\top \succeq c I_{r}\Big\}.
    \end{split}
\end{align}}\normalsize
The exploration threshold $\beta_U(\delta,t)$ is selected to ensure that the algorithm is $(U,\varepsilon,\delta)$-PAC, and is defined in the following lemma.
\begin{lemma}
\label{lem:stopping_rule_eps_delta_pac_passive}
Let $u>0$, and define the exploration threshold
\small{\begin{equation}
    \label{eq:expl_rate_eps_delta_pac_passive}
        \beta_U(\delta, t) = (1+u) \log \left(\frac{\operatorname{det}\left((uc)^{-1} P A_t P^\top + I_{r}\right)^{\frac{1}{2}}}{\delta}\right).
\end{equation}}\normalsize
\noindent Under the stopping rule defined by
\eqref{eq:stopping_rule_eps_delta_pac_passive}--\eqref{eq:expl_rate_eps_delta_pac_passive} and under any sampling rule, we have
     $$\mathbb{P}_{\theta}\left(\exists x\in \mathcal{X}: \theta^\top (\phi_{x,a^\star_\theta(x)} - \phi_{x,\hat{a}_\tau(x)}) >\varepsilon, \tau <\infty\right)\leq\delta.$$
\end{lemma}

\subsection{Sample complexity guarantees} 
Next, we establish performance guarantees for the algorithms presented above. Our algorithms achieve a sample complexity, matching (up to a universal multiplicative constant) the approximated lower bound $T^\star_{\theta,\varepsilon}\mathrm{kl}(\delta,1-\delta)$ (or $T_{\theta,\varepsilon}(\alpha)\mathrm{kl}(\delta,1-\delta)$ in the passive learning setting). 

\begin{theorem}
\label{th:optimality}
In the active learning setting, the algorithm defined by \eqref{eq:recommendation_rule}-\eqref{eq:stopping_rule_eps_delta_pac}-\eqref{eq:sampling_rule_force}-\eqref{eq:sampling_rule_track}
is $(\varepsilon,\delta)-$PAC, and its sample complexity satisfies, $\forall \theta\in \mathbb{R}^d$, $\mathbb{P}_\theta\Big(\limsup_{\delta \to 0} \frac{\tau}{\log \left(\frac{1}{\delta}\right)} \lesssim T_{\theta, \varepsilon}^{\star}\Big) = 1$, and $\limsup_{\delta \to 0} \frac{\mathbb{E}_\theta[\tau]}{\log \left(\frac{1}{\delta}\right)} \lesssim T_{\theta,\varepsilon}^{\star}$.
\end{theorem}

\begin{theorem}
\label{th:optimality_passive}
Let $U$ be a linear subspace of $\mathrm{Span}(\Phi_\alpha)$. The algorithm defined by \eqref{eq:recommendation_rule_passive}-\eqref{eq:stopping_rule_eps_delta_pac_passive}
is $(U,\varepsilon,\delta)-$PAC, and its sample complexity satisfies, $\forall \theta\in U$,
$\mathbb{P}_\theta\Big(\limsup_{\delta \to 0} \frac{\tau}{\log \left(\frac{1}{\delta}\right)} \lesssim T_{\theta,\varepsilon}(\alpha)\Big) = 1$, and $\limsup_{\delta \to 0} \frac{\mathbb{E}_\theta[\tau]}{\log \left(\frac{1}{\delta}\right)} \lesssim T_{\theta,\varepsilon}(\alpha)$.
\end{theorem}
\section{Experiments}\label{sec:experiments}
\subsection{Network settings and experimental set-up}
\label{sec:antenna_tilt_experiments}

\begin{figure}
\centering
  \includegraphics[width = 0.8\columnwidth]{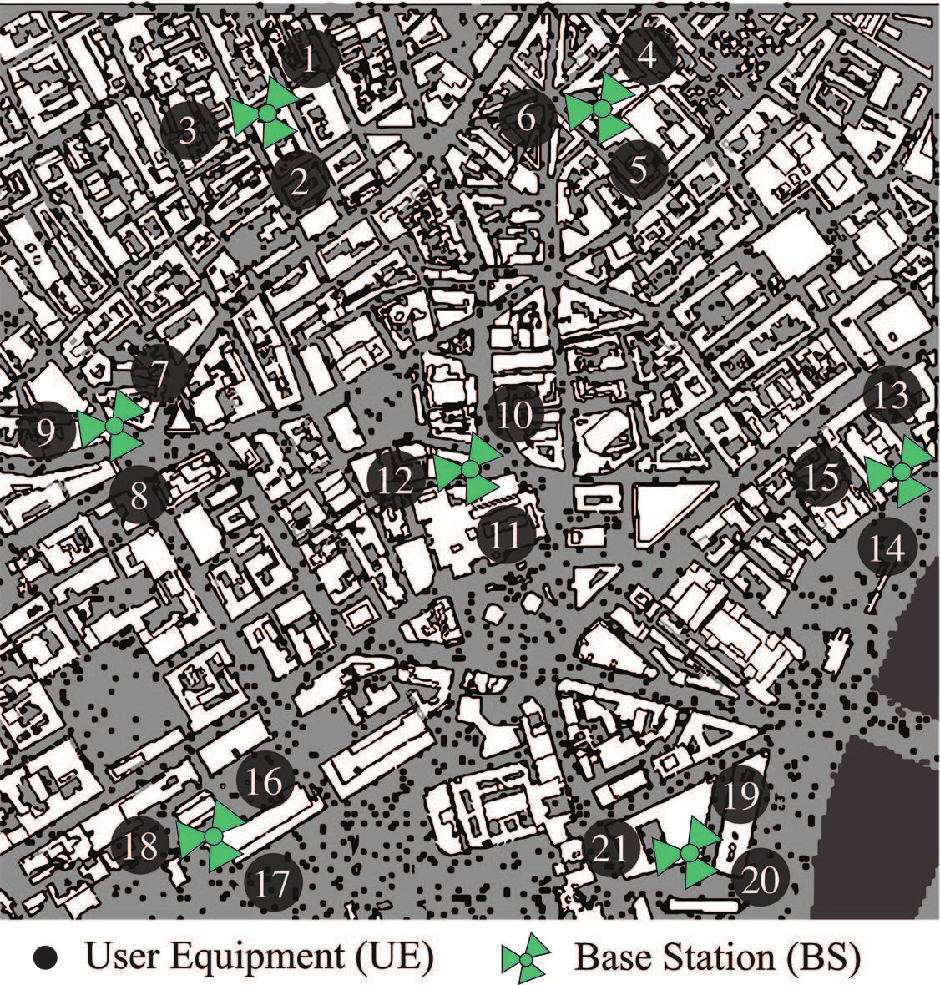}
  \caption{Network environment.}
  \label{fig:network_env}
  \vskip -0.1in
\end{figure}

\noindent
{\it Network simulator.} We run the RET optimization experiments in a 4G simulation environment. The simulation is executed on an urban network consisting of $\mathcal{B}$ BSs, $\mathcal{S}$ sector antennas, and $\mathcal{U}$ users randomly positioned in the environment. The network environment is shown in Fig. \ref{fig:network_env}, and the simulation parameters reported in Table \ref{tab:simulator_setup}. Once the user positions and network parameters are provided, the simulator computes the path loss in the urban environment using the Okomura-Hata propagation model \cite{Rappaport01}, and returns a set of KPIs by conducting user association and resource allocation in a full-buffer traffic demand scenario.

\medskip
\noindent
{\it Learning process.} At each round $t\geq 1$ the network environment generates a new set of random antenna tilts $v_{s,t}$, and user positions, for all $s\in\mathcal{S}$. This yields new values for $N_{\text{OS}}(s,t)$ and $R_{\text{OS}}(s,t)$, that represents the current context (see Sec. \ref{sec:LCB_RET_formulation}). Subsequently, the agent proposes a tilt change $a_{s,t}$, observes $N_{\text{OS}}(s,t+1)$ and $R_{\text{OS}}(s,t+1)$, and computes $p_{s,t}$. The goal is to identify the best tilt change $a_{s,t}$ that, given the current context $x_{s,t}$, produces improved network coverage and capacity as described by the performance function $p_{s,t}$. 

\begin{figure}[ht!]
    \centering
    \includegraphics[width = \columnwidth]{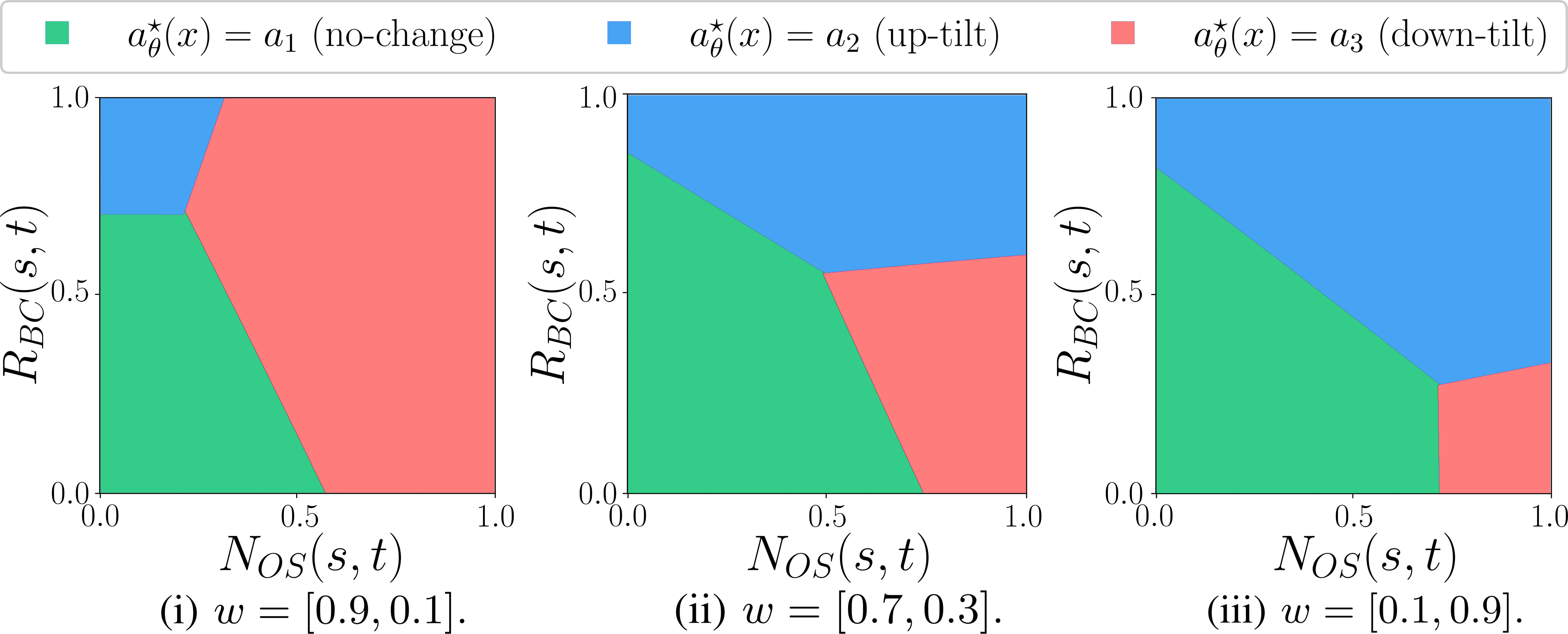}
    \caption{Policy decision regions for different values of $w$.}
    \label{fig:DecisionBoundaries2D}
\end{figure}

\medskip
\noindent
{\it Optimal update policy.} The best tilt update policy depends on the KPIs importance weight vector $w$. We show the impact of $w$ on $p_{s,t}$ in Fig. \ref{fig:DecisionBoundaries2D}. There, we present the policy decision regions when varying $N_{\text{OS}} \left(s,t\right)$ and $R_{\text{BC}}\left(s,t\right)$, for different values of $w$. Note that when both values of $N_{\textsc{OS}}\left(s,t\right)$ and $R_{\text{BC}}\left(s,t\right)$ are low (i.e., there is an acceptable level of coverage and capacity in the sector), the no-change action is optimal. For high values of $N_{\text{OS}}\left(s,t\right)$ (i.e., when there is a problem in the capacity of the sector), the \textit{down-tilt} action is predicted to be the best. On the contrary, for high values of $R_{\textsc{BC}}\left(s, t\right)$, occurring when there is a problem in the capacity of the sector, the \textit{up-tilt} action becomes optimal. It can be observed that, based on the values of $w$, greater importance is given to $R_{\text{BC}}$ or $N_{\text{OS}}$ in determining of the decision regions. 

\begin{table}
\centering
\caption{Simulator parameters.}
\begin{tabular}{lcl}
\toprule
\textsc{Simulator parameter} & \textsc{Symbol} & \textsc{Value} \\ \midrule \midrule 
Number of BSs & $\mathcal{B}$ & $7$ \\
Number of sectors & $\mathcal{S}$& $21$ \\
Number of UEs & $\mathcal{U}$ &$2000$ \\
Carrier frequency & $f$ & $1800$ MHz \\
Antenna height&$h$ & $32$ m \\
Minimum tilt angle & $v_{\text{min}}$ & $1^{\circ}$\\
Maximum tilt angle & $v_{\text{max}}$& $16^{\circ}$ \\ \bottomrule 
\label{tab:simulator_setup}
\end{tabular}
\vskip -0.2in
\end{table}

\medskip
\noindent
{\it Algorithms.} In the passive learning setting, in which the sampling rule is fixed, we implement two types of sampling rules: $(i)$ a \textit{random sampling rule}, i.e., $a_t(x)$ is selected with probability $\alpha_{x,a} = 1/K$, for all $(x,a)\in\mathcal{X}\times\mathcal{A}$, and $(ii)$ a \textit{rule-based sampling rule}, selecting actions according to the fuzzy logic algorithm proposed by Buenostado et al. \cite{Buenostado17}. Note that, as opposed to the random sampling, the rule-based one is deterministic, i.e., $\forall x\in\mathcal{X}$, there exists $a(x) \in\mathcal{A}$ such that $\alpha_{x,a(x)} = 1$, and for all $a\in\mathcal{A}\setminus \{a(x)\}$, $\alpha_{x,a(x)} = 0$. For the passive learning setting, we use the decision rule in \eqref{eq:LS_estimator} and the stopping rule in \eqref{lem:stopping_rule_eps_delta_pac}. For the active learning setting, we use the tracking rule as specified in \eqref{eq:sampling_rule_force}-\eqref{eq:sampling_rule_track}.

The algorithm parameters are fixed as $u = 1$ and $c = 0.1$, and we execute our experiments for $\delta = 0.1$ and $\varepsilon \in \{0.1,0.05,0.025\}$. We use $w = [0.6,0.4]$ in our experiments. Since $\mathcal{X}$ is continuous, we use a discrete uniform mesh having $\mathcal{C} = 20^d$ bins to discretize it. We report the lower bound on the sample complexity in terms of $T^\star_{\theta,\varepsilon}\mathrm{kl}(\delta,1-\delta)$ and the empirical sample complexity in terms of mean and standard deviation over $N_{\text{sim}} = 10^3$ runs. Note that to compute $T^\star_{\theta,\varepsilon}$, we use the parameter $\theta$ obtained fitting observations to a linear model as explained below.
\subsection{Linear model validation}
\label{sec:validation}
\begin{figure}[ht!]
    \centering
    \includegraphics[width = 0.75\columnwidth]{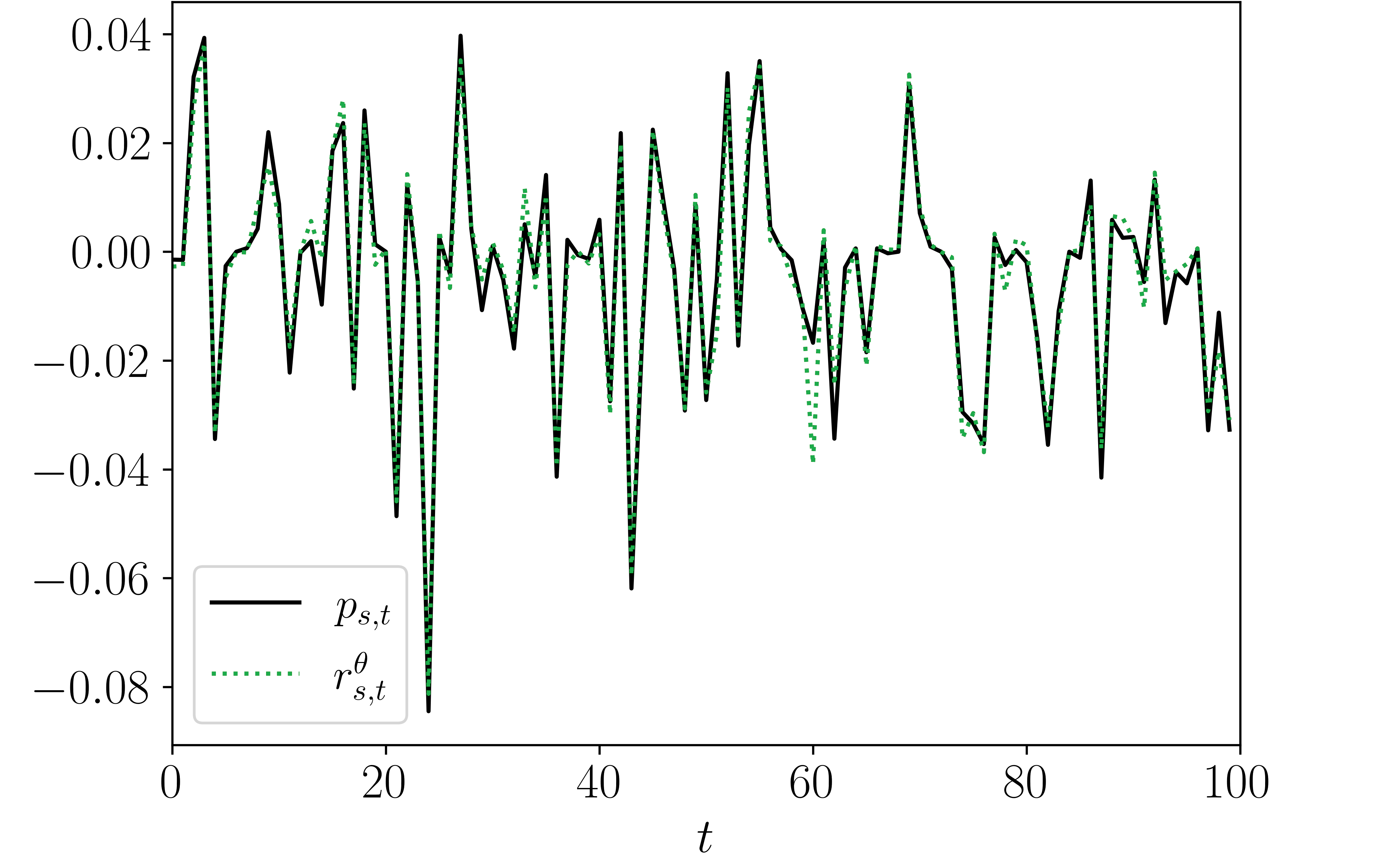}
    \caption{Comparison of $p_{s,t}$ and $r^\theta_{s,t}$ on $D_{\text{test}}$.}
    \label{fig:MSE}
\end{figure}

We present here results validating the linear model used in the paper. To validate the model, we use the dataset $D$, with $N = 1,125,537$ samples, collected while running the BPI algorithms. First, we split $D$ in a training-set $D_{\text{train}}$ ($80\%)$ and a testing-set $D_{\text{test}}$ ($20\%$), containing $N_{\text{train}} = 900,430 $ and $N_{\text{test}} = 225,107$ samples, respectively. Then, we fit a linear model $\theta$ by computing the LSE solution on the samples contained in $D_{\text{train}}$. We measure the Normalized Root Mean-Squared Error (NRMSE) on $D_{\text{test}}$ between the performance function $p_{s,t}$ (see Section \ref{sec:LCB_RET_formulation}) and the fitted reward model $r^\theta_{s,t} = \theta^\top\phi_{x_{s,t},a_{s,t}}$. The NRMSE is defined as $\mathrm{NRMSE}(\theta) = \frac{1}{\bar{p}}\sqrt{\frac{1}{\mathcal{S}N_\text{test}}\sum_{s = 1}^\mathcal{S}\sum_{t = 1}^{N_{\text{test}}} (p_{s,t} - r^{\theta}_{s,t})^2}$, where $\Bar{p} = \max_{s,t} p_{s,t} - \min_{s,t} p_{s,t}$ is the range of $p_{s,t}$. We attain an NRMSE equal to $\mathrm{NRMSE}(\theta) = 0.15$. Fig. \ref{fig:MSE} reports a sample of the fit between $p_{s,t}$ and $r^{\theta}_{s,t}$ from $100$ randomly sampled data points from $D_{\text{test}}$. The fit is extremely accurate. 

\subsection{Results}
The experimental results are shown in Table \ref{tab:results}. Before commenting these results, we note that both sampling rules (random and rule-based) in the passive learning setting allow us to determine $\theta$ accurately, i.e., $U=\mathbb{R}^d$.  

A first observation is that as expected, an active learning approach yields lower sample complexity than the passive learning approach. This can be seen for both the lower bounds and the actual sample complexity achieved under our algorithms. Indeed, being active allows us to explore unexplored dimensions of $\theta$ and in turn to learn faster.

Further, we observe that collecting data using a random policy yields better results than using the rule-based policy. This can be easily explained by noting that the random policy explores more than the rule-based policy and hence speeds up the learning process.

Finally, we note that the proposed algorithms exhibit a sample complexity close to the corresponding information-theoretical limit (they differ from a small multiplicative factor depending on the setting). 

\begin{table}[ht!]
\centering
\caption{BPI results for RET optimization.}
\begin{tabular}{clll}\toprule
Sampling rule & $\varepsilon$ & $T^\star_{\theta,\varepsilon}$ & Sample complexity \\ \midrule \midrule
\multirow{3}{*}{Rule-Based} 
& $0.10$  & $123.58$ & $246.16\pm 64.23$  \\ & $0.05$ & $425.63$ & $719.01\pm142.93$  \\ & $0.025$ & $1531.47$ & $2384.03 \pm 313.01$  \\
\midrule 
\multirow{3}{*}{Random} 
& $0.10$  & $69.40$ & $240.58\pm92.75$  \\ & $0.05$  & $231.69$ & $494.55 \pm 89.89$ \\ & $0.025$ & $1487.28$ & $1857.50\pm332.48$  \\ \midrule
\multirow{3}{*}{Adaptive} 
& $0.10$  & $53.23$ & $226.32 \pm52.73$  \\ & $0.05$  & $214.83$ & $444.51 \pm 49.94$ \\ & $0.025$ & $1286.38$ & 
$1545.24\pm212.93$ \\ \bottomrule
\end{tabular} 
\label{tab:results}
\vskip -0.15in
\end{table}

\section{Proofs}\label{sec:lemmas}
Due to space constraints, we focus on the active learning setting. The results for passive learning can be established in a similar way. The proofs of many intermediate results are similar to those of the corresponding results for plain linear bandits presented in \cite{Jedra20}. The main difference lies in the fact that, since the sequence of contexts is random (not under our control), we can obtain results with probabilistic guarantees only. This is for example the case for the growth rate of the smallest eigenvalue of the covariates matrix (we could have deterministic guarantees in absence of contexts \cite{Jedra20}). To deal with this randomness, we present the following lemma that provides a concentration result on the number of times a context is observed, and that will be extensively applied in the proofs of the remaining results. 
\begin{lemma}
\label{lem:Nx_concentration_tight}
For any $t\geq 1$, and $\varepsilon>0$, we have
$$\mathbb{P}_\theta\left(\max_{x\in\mathcal{X}} \left|\frac{N_x(t)}{t}-p_{\mathcal{X}}(x)\right|\geq \varepsilon\right) \leq 2e^{-\frac{t \varepsilon^2}{4}}.$$
\end{lemma}
\subsection{Sketch of the proof of Lemma \ref{lem:Nx_concentration_tight}}
The proof follows by applying the concentration result on vector-valued martingales in \cite[Th. 3.5]{iosif_concentration}. Define $N_t-tp_\mathcal{X}=\sum_{s = 1}^t Y_s,$
where $N_t$ is the random vector in $\mathbb{N}^{|\mathcal{X}|}$ with coordinates $N_x(t)$ for $x\in\mathcal{X}$, $p_\mathcal{X}$ is the vector with coordinates $p_\mathcal{X}(x)$ that lies in $\Lambda_\mathcal{X} = \{p_{\mathcal{X}}(x) \in[0,1]^{|\mathcal{X}|}: \sum_{x\in\mathcal{X}} p_\mathcal{X}(x) = 1\}$, and 
$Y_s$ is an i.i.d. zero-mean random vector in $\mathbb R^{|\mathcal{X}|}$ with coordinates $Y_s(x)=\mathbbm{1}_{\{x_s=x\}} - p_\mathcal{X}(x)$ for $x\in \mathcal{X}$. We have 
\begin{align*}
   \|Y_s\|_2^2 & =\sum_{x\in\mathcal{X}}(\mathbbm{1}_{\{x_s=x\}}-p_\mathcal{X}(x))^2 \\&\le \sum_{x\in\mathcal{X
}}(\mathbbm{1}_{\{x_s=x\}}+p_\mathcal{X}(x)^2)  = 1+\sum_{x\in\mathcal{X}}p_\mathcal{X}(x)^2\le2. 
\end{align*}
Hence the assumptions of \cite[Th. 3.5]{iosif_concentration} are satisfied with $r=t\varepsilon$, $D=1$, and $b_*^2=2t$, and we get
{\small\begin{align*}
& \mathbb{P}\Bigg( \max_{x\in\mathcal{X}}\left|\frac{N_x(t)}{t}-p_\mathcal{X}(x)\right|\ge\varepsilon\Bigg) = \mathbb{P}\bigg(\max_{x\in\mathcal{X}}\left|N_x(t)-tp_{\mathcal{X}}(x)\right|\ge t\varepsilon\bigg) \\
& = \mathbb{P}\left(\|N_t-t p_\mathcal{X}\|_\infty \geq t\varepsilon\right) 
\leq \mathbb{P}\left(\|N_t-t p_\mathcal{X}\|_2 \geq t\varepsilon\right) \leq 2 e^{-t\varepsilon^2/4}. \end{align*}}\normalsize 

\subsection{Sketch of the proof of Lemma \ref{lem:stopping_rule_eps_delta_pac}}
The main step of the proof is to show that the probability of the error event, in which the stopping condition is satisfied and there is a context for which the algorithm outputs an $\varepsilon$-suboptimal arm, is smaller than $\delta$.
This follows by $(i)$ upper bounding $Z^x_\varepsilon(t)$ in terms of $\hat{\theta}_t$ and $(ii)$ applying the concentration results for self-normalized processes by \cite{Abbasi11}. More precisely, one can show that for all $x\in\mathcal{X}$,
$Z^x_\varepsilon(t) \leq Z^x_{\hat{a}_t(x), a^\star_\theta(x), \varepsilon}(t) \leq \frac{1}{2}\|\theta - \hat{\theta}_t\|_{A_t}^2.$ Also, it is easy to verify that, under the error event $\mathcal{E} = \Big\{\exists t, \exists x: Z_\varepsilon^x(t)>\beta(\delta, t)\text{ and }  \theta^\top(\phi_{x,a^\star_\theta(x)} - \phi_{x,\hat{a}_t(x)})>\varepsilon  \text{ and } A_t \succeq c I_{d} \Big\}$, and for $u>0$, we have
$$
\|\theta-\hat{\theta}_{t}\|_{A_t}^{2} \leq(1+u)\left\|\sum_{s=1}^{t} \phi_{x_s,a_s} \xi_{s}\right\|_{\left(A_t+u c I_{d}\right)^{-1}}^{2}.
$$
Hence, the error event satisfies 
{\small\begin{align*}
    \mathcal{E} \subseteq \left\{\exists t ,\exists x:  \frac{1+u}{2}\left\|\sum_{s=1}^{t} \phi_{x_s,a_s} \xi_{s}\right\|_{\left(A_t^{\top}+u c I_{d}\right)^{-1}}^{2} \geq \beta(\delta, t)\right\},
\end{align*}}\normalsize
The result follows immediately by selecting $\beta(\delta,t)$ as in  \eqref{eq:expl_rate_eps_delta_pac} and applying the concentration results in \cite[Theorem 1]{Abbasi11}.

\subsection{Sketch of the proof of Lemma \ref{lem:forced_exploration}}
The proof proceeds as follows. First, we show that if for some context $x\in\mathcal{X}$, the condition $A_t(x) \succ f_x(t)$ is violated, then the number of contexts to be observed in order to satisfy the condition again cannot exceed $d_x$ rounds. In fact, one can actually show using similar arguments to \cite[Lem. 5]{Jedra20} applied to each context, to show that exists $t_0 \geq 1$, such that for all $t\geq t_0$, and for all $x\in\mathcal{X}$, $A_t(x) \succeq C_x\sqrt{N_x(t)-d_x-1}$, where $C_x = \frac{1}{\sqrt{d_x}}\sum_{a_x \in \mathcal{A}_x} \phi_{x,a_x} \phi_{x,a_x}^{\top}$. 
Then, applying the concentration results of Lemma \ref{lem:Nx_concentration_tight}, we have $\forall \varepsilon \in\left(0,p_{\min}\right)$, $\forall t\geq t_0$, and w.p. at least $1-2 e^{-\frac{t\varepsilon^2}{4}}$, $A_t(x) \succeq  C_x\sqrt{(p_\mathcal{X}(x)-\varepsilon)t - d_x - 1}.$ By summing over $x$, one can show that
$$
A_t = \sum_{x \in \mathcal{X}}A_t(x) \succeq \min_{x\in\mathcal{X}}\sqrt{\frac{(p_\mathcal{X}(x)-\varepsilon)t-(d_x+1)}{d_x}}  cI_d,
$$
and the result follows by determining, from the above expression, a $t_1(\kappa) \geq t_0$ such that $A_t \succeq \kappa \sqrt{t} I_d.$

\subsection{Sketch of the proof of Lemma \ref{lem:tracking_lemma}}
The proof follows similar steps to \cite[Lem. 6]{Jedra20}. Such steps are applied to each of the contexts, and one can finish by applying the concentration results in Lemma \ref{lem:Nx_concentration_tight}. A key difference is that we actually require convexity on each of the sets $C_x \subset C$, that is guaranteed by the convexity of $C$. To see this, define for $x\in\mathcal{X}$, the projection operator $\text{proj}_x(\cdot)$ that takes $\alpha \in C$ as input, and outputs $\alpha_x$, the $K$-dimensional vector containing the component of $\alpha$ relative to the context $x$. Let $C_x = \{\alpha_x: \alpha_x = \text{proj}_x(\alpha), \forall \alpha \in C\}$.
Note that, since $C \subset \Lambda$ is convex, we have that each of the sets $C_x$ is also convex, and we have $   \forall t\geq t_0, \forall x\in\mathcal{X}, d_\infty(\alpha_x, C_x) \leq \varepsilon.
$
Now, define for all $t\geq 1$, $\varepsilon_{x,a,t} = N_{x,a}(t) - N_x(t)\hat{\alpha}_{x,a}(t)$. One can prove using similar arguments to \cite[Lem. 6]{Jedra20} that (i) there exists $t_0^{\prime}\geq t_0$, $c>0$ such that for all $t \geq t_0'$ we have $\varepsilon_{x,a,t}\leq\left(z_{t}+d- 1\right) \max \left\{t_{0}^{\prime}, c+1\right\},$ and (ii)
for $t\geq t_0'$, $
d_{\infty}\left(\left(N_{x,a}(t)/N_{x}(t)\right)_{(x,a) \in \mathcal{X}\times \mathcal{A}}, C\right) 
\leq \max _{(x,a) \in \mathcal{X}\times \mathcal{A}}\left|\frac{\varepsilon_{x,a,t}}{N_x(t)}\right|. $
Finally, applying Lemma \ref{lem:Nx_concentration_tight}, and determining $t_1(\varepsilon,u)$ such that such that for all $t \ge t_1$ we have $\max_{(x,a) \in \mathcal{X}\times \mathcal{A}}\left|\frac{\varepsilon_{x,a,t}}{(p_\mathcal{X}(x) -u)t}\right| \leq (z_t + d -1)\varepsilon$,  the result follows.

\subsection{Sketch of the proof of Theorem \ref{th:optimality}}
By Lemma \ref{lem:LS_concentration}, we have $\lim_{t\to \infty} \theta_t = \theta$, a.s. and by Lemma \ref{lem:convergence_proportions} we have $\lim_{t\to \infty}d_\infty\left(\left(N_{x,a}(t)/N_x(t)\right)_{(x,a)\in\mathcal{X}\times\mathcal{A}},C^\star_\varepsilon(\theta) \right) = 0$ a.s. Let $q>0$. By continuity of $\bar{\psi}_\varepsilon$ (Lemma \ref{lem:properties_eps_delta}), one can actually show that there exists $t_0 \geq 1$ such that for all $t\geq t_0$, $ \bar{\psi}_\varepsilon\left(\hat{\theta}_t,\left(N_{x,a}(t)/N_x(t)\right)_{(x,a)\in\mathcal{X} \times \mathcal{A}}\right) \geq (1-q)\bar{\psi}_\varepsilon\left(\theta,\alpha^\star \right),$ for some $\alpha^\star \in C^\star_\varepsilon(\theta)$.
By the definitions of $Z^x_{\varepsilon}$ and $\bar{\psi}_\varepsilon$, for all $t\geq t_0$ we have\\
$
\min_{x\in\mathcal{X}} Z^x_\varepsilon(t) \geq  t \bar{\psi}_\varepsilon\left(\hat{\theta}_{t},\left(N_{x,a}(t)/N_x(t)\right)_{(x,a)\in\mathcal{X} \times \mathcal{A}}\right).
$
By Lemma \ref{lem:forced_exploration}, and by applying the first Borel-Cantelli Lemma, there exists a $t_1 \geq 1$ such that $A_t \succeq c I_d$, for some $c > 0$. This implies that for all $t \geq \max \{t_0,t_1\}$ we have 
\small{\begin{align*}
& \tau  =\inf \Big\{t: \min_{x\in\mathcal{X}} Z_\varepsilon^x(t)>\beta(\delta, t) \text{ and } A_t \succeq c I_{d}\Big\} \\&  \leq \max \{t_0,t_1\} \lor \inf \Big\{t: t(1-q) \bar{\psi}_\varepsilon\left(\theta,\alpha^\star \right)>\beta(\delta, t),\ A_t \succeq c I_{d}\Big\} \\ & \lesssim \max\left\{t_0,t_1, \frac{1}{1-q}T^\star_{\varepsilon,\theta}\log\left(\frac{1}{\delta}\right)\right\},
\end{align*}}\normalsize
where we used \cite[Lem. 18]{Garivier16} for $\delta$ sufficiently small in the last inequality. When $q\to 0$, we get 
$
\mathbb{P}_\theta(\limsup_{\delta \to 0} \frac{\tau}{\log (\frac{1}{\delta})} \lesssim T_{\varepsilon,
\theta}^{\star}) = 1.
$
The proof of the guarantee in expectation is more complex. It proceeds using similar arguments as \cite[Th. 3]{Jedra20}. First, we construct an event $\mathcal{E}_T$ under which, for all $t\geq T$, the stopping time is well-behaved. The event includes the probabilistic results for the tracking and forced exploration rules in Lemma \ref{lem:forced_exploration} and \ref{lem:tracking_lemma}. One can then show that the probabilities of the events under which the sample complexity is not well-behaved are negligible. The proof is concluded by giving an upper bound on the expected sample complexity.

\section{Conclusions} 
\label{sec:conclusions}
In this paper, we solved the BPI problem in linear contextual bandits: we derived sample complexity lower bounds and devised algorithms achieving these limits in both the active and passive learning settings. We applied the CL-MAB framework to the RET optimization problem, and demonstrated that our algorithms can identify the best tilt update policy using much fewer samples than rule-based or naive algorithms. Our results are promising and could be extended to various other optimization problems, such as the design of BPI algorithms that accounts for coordination across different sectors. We also plan to investigate, using our framework, the problem of beam management in multiple antenna systems such as Multiple Input Multiple Output (MIMO) systems in 5G networks. 

\section*{Acknowledgements}
This work was partially supported by the Wallenberg AI, Autonomous Systems and Software Program (WASP) funded
by the Knut and Alice Wallenberg Foundation.


\bibliography{biblio.bib}
\bibliographystyle{plainnat}
\end{document}